\documentclass[runningheads,a4paper]{llncs}

\usepackage{graphicx}        

\usepackage{todonotes}
\usepackage{algorithm,algorithmicx}
\usepackage{algpseudocode}
\usepackage{amssymb}
\usepackage{url}
\urldef{\mailsa}\path|{madalina.olteanu,nathalie.villa,marie.cottrell}@univ-paris1.fr|

\begin{document}

\mainmatter  

\title{On-line relational SOM for dissimilarity data}

\titlerunning{On-line relational SOM for dissimilarity data}

%
%
\author{Madalina Olteanu%
\and Nathalie Villa-Vialaneix
\and Marie Cottrell}
\authorrunning{On-line SOM for dissimilarity data}

\institute{SAMM-Universit\'e Paris 1 Panth\'eon Sorbonne\\
90, rue de Tolbiac, 75013 Paris - France\\
\mailsa\\
\url{http://samm.univ-paris1.fr}}

%
%

\toctitle{On-line SOM for dissimilarity data}
\tocauthor{On-line SOM for dissimilarity data}
\maketitle

\begin{abstract}
In some applications and in order to address real world situations better, data may be more complex 
than simple vectors. In some examples, they can be known through their pairwise dissimilarities only.
Several variants of the Self Organizing Map algorithm were introduced to generalize the
original algorithm to this framework.
 Whereas median SOM is based on a rough representation of the
prototypes, relational SOM allows representing these prototypes by a virtual
combination of all elements in the data set. However, this latter approach suffers
from two main drawbacks. First, its complexity can be large. Second, only a batch
version of this algorithm has been studied so far and it often
provides results having a bad topographic organization. In this article, an
on-line version of relational SOM is described and justified. The algorithm
is tested on several datasets, including categorical data and graphs, and compared
with the batch version and with other SOM algorithms for non vector data.
\end{abstract}

\section{Introduction}

In many real-world applications, data cannot be described by a fixed set of
numerical attributes. This is the case, for instance, when data are
described by categorical variables or by relations between objects (i.e., persons involved in
a social network). A common solution to address this kind of issue is to use a measure of
resemblance (i.e., a similarity or a dissimilarity) that can handle categorical variables,
graphs or focus on specific aspects of the data, designed by expertise
knowledge. Many standard methods for data mining have been generalized to non
vectorial data, recently including prototype-based clustering. The
recent paper \cite{cottrell_etal_KI2012} provides an overview of several methods
that have been proposed to tackle complex data with neural networks.

 In
particular, several extensions of the Self-Organizing Maps (SOM) algorithm have
been proposed. One approach consists in extending SOM to categorical data by using a method
similar to Multiple Correspondence Analysis, \cite{cottrell_letremy_N2005}.
Another approach uses the median principle which consists in replacing the
standard computation of the prototypes by an approximation in the
original dataset. This principle was used to extend SOM to dissimilarity data in
\cite{kohonen_somervuo_N1998}. One of the main drawbacks of this approach is
that forcing the prototypes to be chosen among the dataset is very
restrictive; in order to increase the
flexibility of the representation, \cite{conanguez_rossi_elgolli_NN2006} propose to represent a class
by several prototypes, all chosen among the original dataset. However this
method increases the computational time and prototypes 
still stay restricted to the original dataset, hence reflecting possible
sampling or sparsity issues.

An alternative to median-based algorithms relies on a method that is
close to the classical algorithm used in the Euclidean case and is based on the idea that prototypes may be expressed as linear
combinations of the original dataset. In the kernel SOM framework, this setting
is made natural by the use of the kernel that maps the original data into a
(large dimensional) Euclidean space (see
\cite{macdonald_fyfe_ICKIESAT2000,andras_IJNS2002} for on-line versions and
\cite{boulet_etal_N2008} for the batch version). Many kernels have been
designed to handle complex data such as strings, nodes in a graphs or graphs
themselves \cite{gartner_KSD2008}. 

More generally, when the data are already described by a
dissimilarity that is not associated to a kernel,
\cite{hammer_etal_WSOM2007,rossi_etal_WSOM2007,hammer_etal_WSOM2011} use a
similar idea. They  introduce
an implicit ``convex combination'' of the original data to extend the classical batch
versions of SOM to dissimilarity data. This approach
is known under the name ``relational SOM''. The purpose of the present paper is to show that the same idea can be used to
define on-line relational SOM. Such an approach reduces the computational cost of the
algorithm and leads to a better organization of the map. In the remaining of
this article, Section~\ref{method} describes the methodology and
Section~\ref{applications} illustrates its use on 
simulated and real-world data.

\section{Methodology}\label{method}

In the following, let us suppose that $n$ input data, $x_1$, \ldots, $x_n$, from an arbitrary input
space $\mathcal{G}$ are given. These data are described by a dissimilarity
matrix $\mathbf{D}=(\delta_{ij})_{i,j=1,\ldots,n}$ such that $D$ is
non negative ($\delta_{ij}\geq 0$), symmetric ($\delta_{ij}=\delta_{ji}$) and 
null on the diagonal ($\delta_{ii}=0$). The purpose of the algorithm is to map
these data into a low dimensional grid composed of $U$ units
which are linked together by a neighborhood relationship $K(u,u')$. 
A prototype $p_u$ is associated with each unit $u\in\{1,\ldots,U\}$ in the grid.
The $U$ prototypes $(p_1, p_2, \ldots,p_U)$ are initialized either randomly
among the input
data
or as random convex combinations of the input data.

\textbf{In the Euclidean framework}, where the input space is equipped with a distance, 
the matrix $D$ is the distance matrix with entries $\delta_{ij}=\|x_i-x_j\|^2$.
In this case, the on-line SOM algorithm iterates
\begin{itemize}
	\item an \emph{assignment step}:  a randomly chosen input $x_i$ is assigned to
the closest prototype denoted by $p_{f(x_i)}$ according to shortest distance rule
\[
	f(x_i) = \arg\min_{u=1,\ldots,U} \|x_i-p_u\|,
\]
	\item a \emph{representation step}: all prototypes are updated
\[
	p_u^{\textrm{new}} = p_u^{\textrm{old}} + \alpha
K(f(x_i),u)\left(x_i-p_u\right), 
\]
where $\alpha$ is the training parameter.
\end{itemize}

\textbf{In the more general framework}, where the data are known through pairwise distances only,
the assignment step cannot be carried out straightforwardly since the distances between the input data and the
prototypes may not be directly computable. The solution introduced in
\cite{rossi_etal_WSOM2007}
consists in supposing that prototypes are convex
combinations of the original data, $p_u=\sum_i \beta_{ui} x_i$ with $\beta_{ui}>0$ and $\sum_i
\beta_{ui}=1$. If $\beta_u$ denotes the vector $(\beta_{u1}, \beta_{u2}, \ldots,
\beta_{un})$, the distances in the assignment step can be written in terms of
$D$ and $\beta_u$ only:
\[
	\|x_i-p_u\|^2 = \left(D\beta_u\right)_i - \frac{1}{2} \beta_u^T D \beta_u.
\]

According to \cite{rossi_etal_WSOM2007}, the equation above still holds if the matrix $D$ is no longer
a distance matrix, but a general dissimilarity matrix, as long as it is
symmetric and null on the diagonal. 
A generalization of the batch SOM algorithm, called batch relational SOM, which holds for 
dissimilarity matrices is introduced in \cite{rossi_etal_WSOM2007}. 

The representation step may also be carried out in this general framework
as long as the prototypes are supposed to be convex combinations of the input data. Hence, using the same
ideas as  \cite{rossi_etal_WSOM2007}, we introduce the on-line relational SOM, which generalizes the 
on-line SOM to dissimilarity data. The proposed algorithm is the following:



\begin{algorithm}
	\caption{On-line relational SOM}
	\label{algo::online-d-som}
	\begin{algorithmic}[1]
		\State For all $u=1,\ldots,U$ and $i=1,\ldots,n,$ initialize $\beta^0_{ui}$
randomly in $\mathbb{R}$, such that $\beta^0_{ui}\ge0$ and $\sum_i^n\beta^0_{ui}=1$.
		\For{t=1,\ldots,T}
                        \State Randomly chose an input $x_i$
			\State \emph{Assignment} : find the unit of the closest prototype
			\[
				f^t(x_i) \leftarrow \arg\min_{u=1,\ldots,U}
				\left(\beta_u^{t-1}\mathbf{D}\right)_i -
				\frac{1}{2}\beta_u^{t-1}\mathbf{D}(\beta_u^{t-1})^T
			\]
			\State \emph{Update of the prototypes}: $\forall\,u=1,\ldots,U$,
			\[
				\beta_u^t \leftarrow \beta_u^{t-1}+\alpha^t K^t(f^t(x_i),u)
\left(\mathbf{1}_i - \beta_u^{t-1}\right)
			\]
			where $\mathbf{1}_i$ is a vector with a single non null coefficient at
the $i${\it th} position, equal to one.
		\EndFor
	\end{algorithmic}
\end{algorithm}

In the applications of Section~\ref{applications}, the parameters of the
algorithm are chosen according to \cite{cottrell98}: the neighborhood $K^t$
decreases in a piecewise linear way, starting
from a neighborhood which corresponds to the whole grid up to a neighborhood
restricted to the neuron itself; $\alpha^t$ vanishes at the rate of $1/t$. Let us remark that if the dissimilarity matrix is a
 Euclidean distance matrix,
relational on-line SOM is equivalent
to the classical on-line SOM algorithm, as long as the $n$ input data contain a
basis of the input space $\mathcal{G}$.

As explained in \cite{fort:advantages}, although batch SOM possesses
the nice properties
of being deterministic and of usually converging in a few iterations, it has several drawbacks such as
bad organization, bad visualization, unbalanced classes and strong dependence on the initialization.
Moreover, the computational complexity of the online algorithm may be significantly reduced with respect to the
batch algorithm. For one iteration, the complexity of the batch algorithm is $\mathcal{O}(Un^3+Un^2)$,
while for the online algorithm it is $\mathcal{O}(Un^2+Un)$. However, since the online algorithm has to scan
all input data, the number of iterations is significantly larger than in the batch case. To summarize,
if $T_1$ is the number of iterations for batch relational SOM and $T_2$ is the number of iterations for online
relational SOM, the ratio between the two computation times will be $T_1n/T_2$.

\begin{figure}[h!]
	\centering
	\begin{tabular}{cc}
		Batch relational SOM (20 iterations) \\
   \fbox{\includegraphics[width=0.6\linewidth]{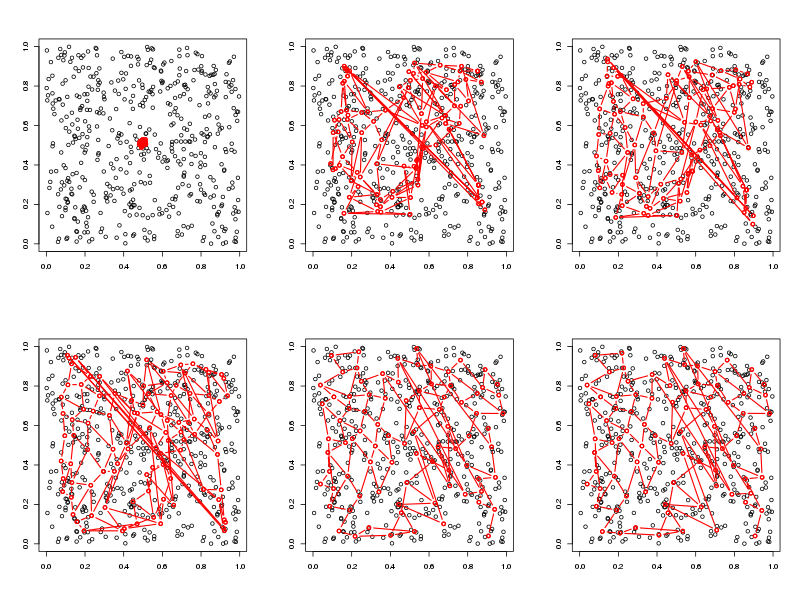}} \\
          On-line relational SOM (2500 iterations) \\
\fbox{\includegraphics[width=0.60\linewidth]{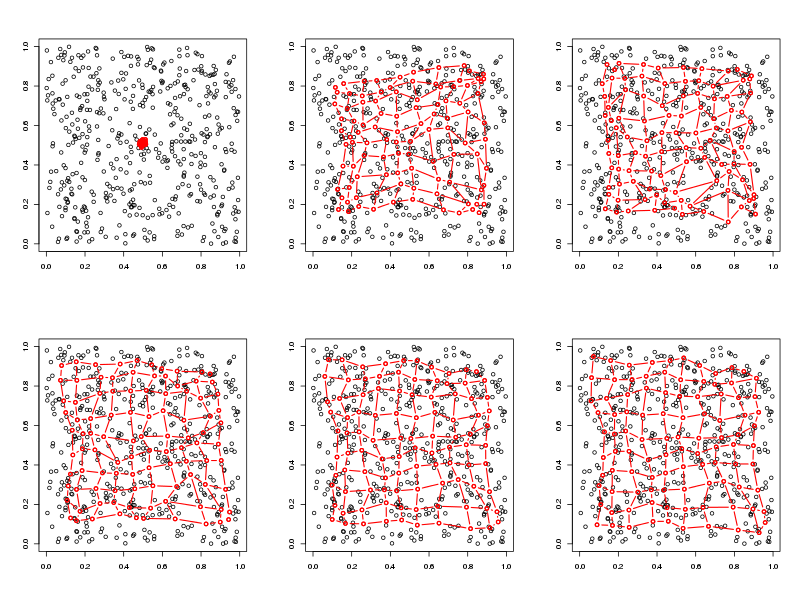}}
	\end{tabular}
	\caption{Batch and on-line SOM organization for 500 samples from the uniform
distribution in 
$\left[0,1\right]^2$.
The same initialization was used for both algorithms.}
	\label{fig::uniform}
\end{figure}

For illustration, let us consider 500 points sampled randomly from the uniform distribution in  $\left[0,1\right]^2$.
The batch version of relational SOM and the on-line version of relational SOM
were
performed with identical 10x10 grid
structures and identical initializations. Results are available in
Figure~\ref{fig::uniform}. Batch relational SOM converged quickly, in 20
iterations (the grid organization is represented at iterations 0
(random initialization), 5, 9, 13,
17 and 20), but the map is not well organized. On-line relational SOM converged
in less than 2500 iterations
(the grid organization is represented at iterations 0 (initialization), 500,
1000, 1500, 2000 and 2500), but the map is now almost
perfectly organized. This results was achieved in 40 minutes for
the batch version and in 10 minutes for the on-line version on a netpc (with
$2\times1$GHz AMD processors and 4Go RAM).

\section{Applications}\label{applications}

This section presents several applications of the on-line relational SOM on various datasets.
Section~\ref{swiss} deals with simulated data described by numerical variables, but
 organized on a non linear
surface. Section~\ref{astraptes} is an application on a real dataset where the
individuals are described by categorical variables. Finally, Section~\ref{polbooks} is an
application to the clustering of nodes of a graph.

\subsection{Swiss roll}\label{swiss}

\begin{figure}[h!]
	\centering
	\begin{tabular}{p{6cm}p{6cm}}
    \fbox{\includegraphics[width=0.9\linewidth]{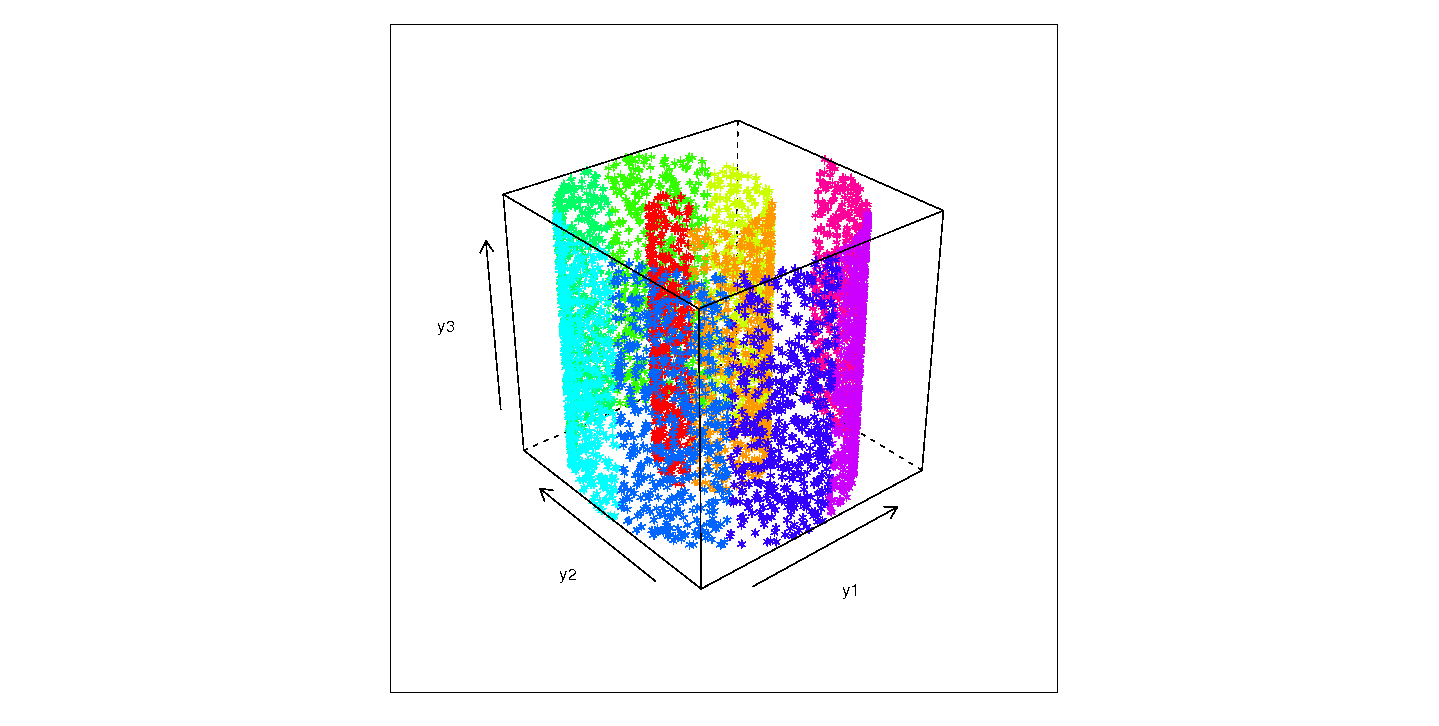}} & \fbox{\includegraphics[width=0.9\linewidth]{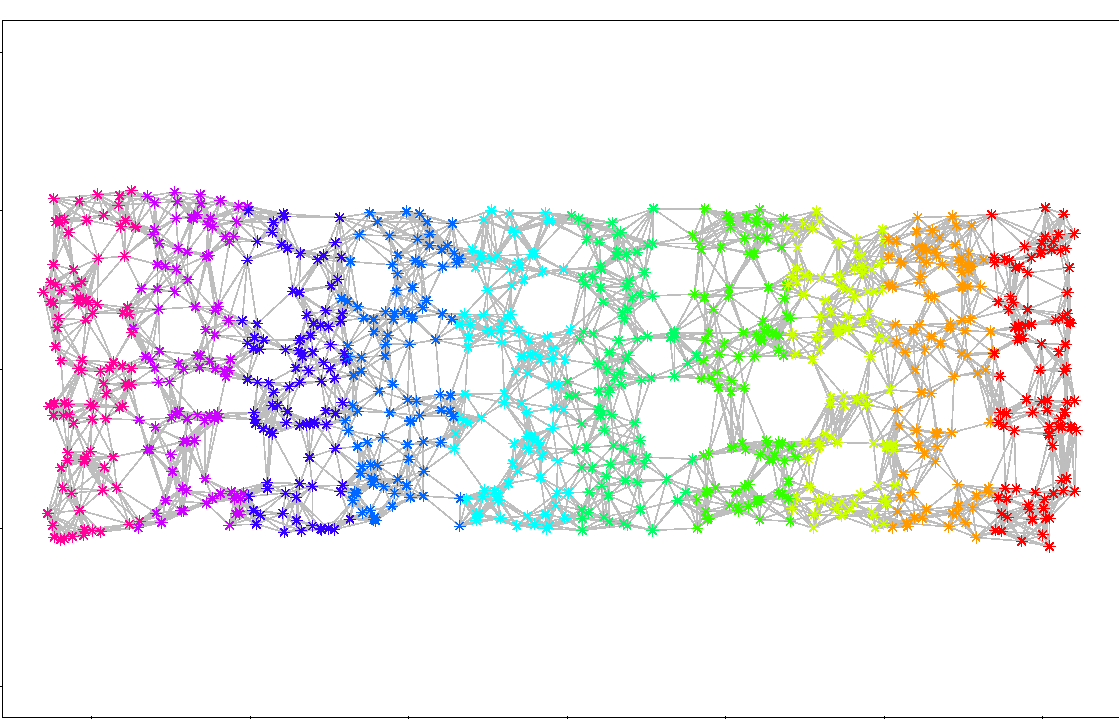}}  \\
	\end{tabular}
	\caption{Unfolding the Swiss roll using Isomap}
	\label{fig::swiss}
\end{figure}

\begin{figure}[h!]
	\centering
	\begin{tabular}{p{6cm}p{6cm}}
     a) 15x15-grid batch median SOM & b) 15x15-grid on-line relational SOM  \\
      \fbox{\includegraphics[height=3cm,width=0.9\linewidth]{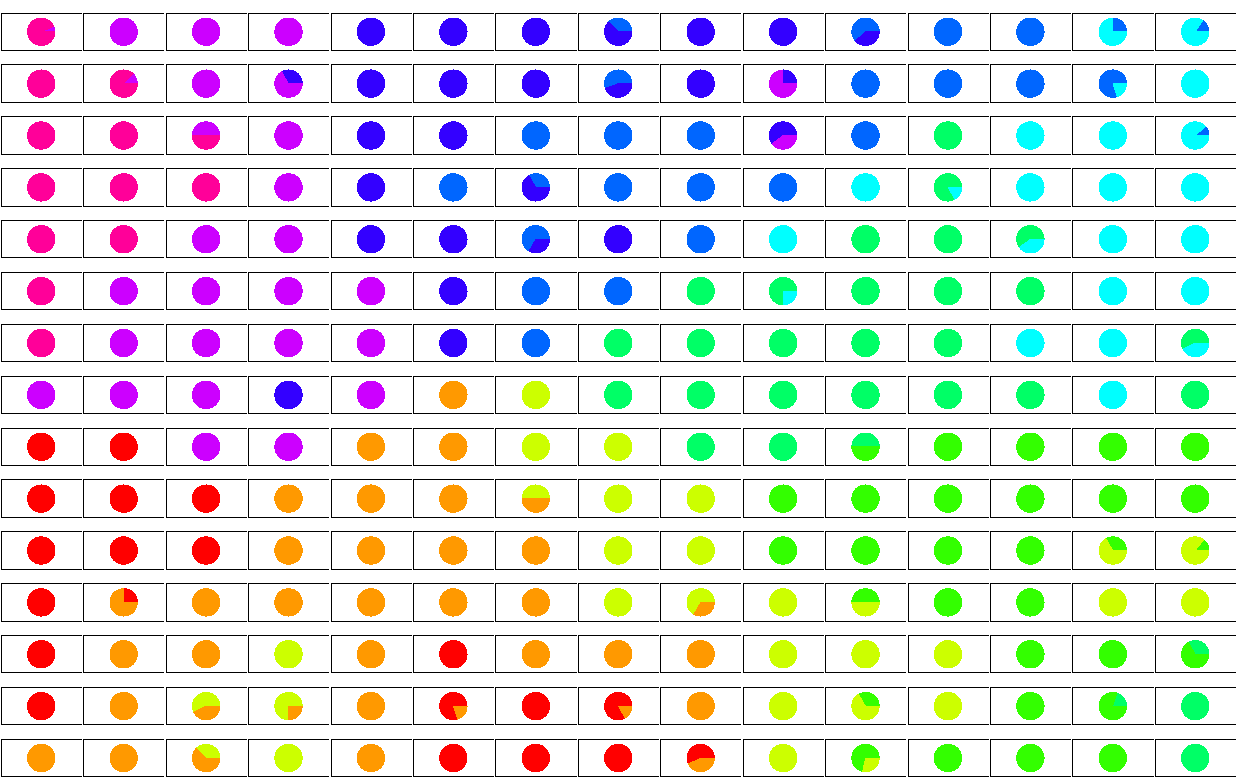}} 
    &  \fbox{\includegraphics[height=3cm,width=0.9\linewidth]{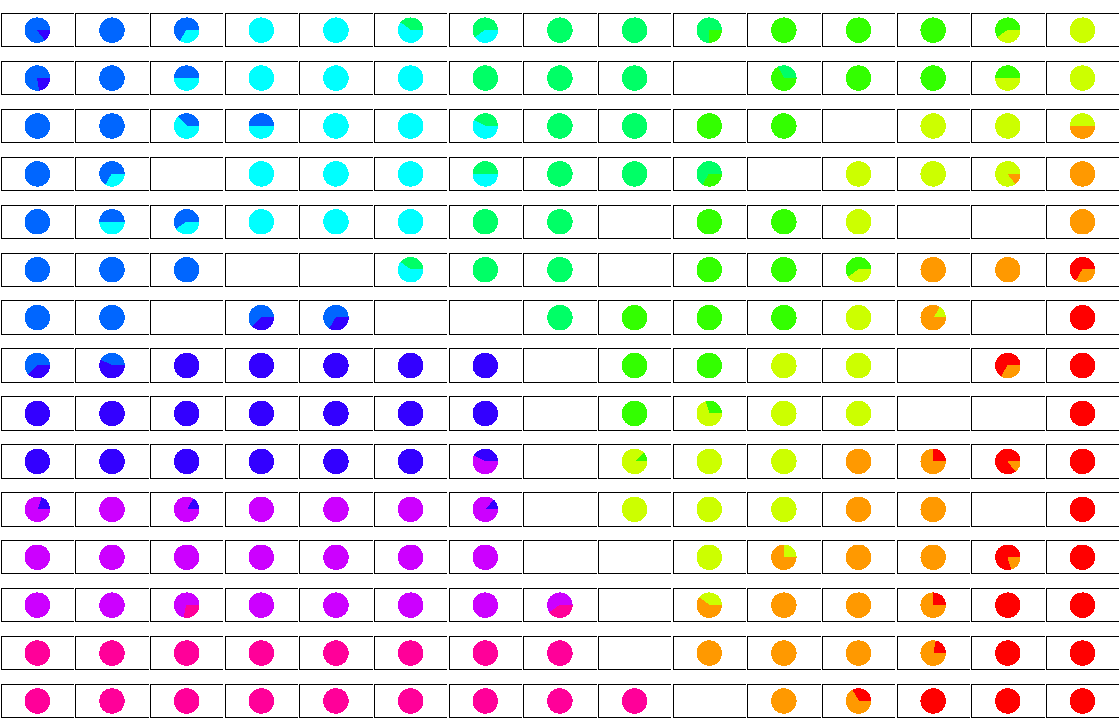}} \\
     c) 30x10-grid batch median SOM & b) 30x10-grid on-line relational SOM  \\
      \fbox{\includegraphics[height=3cm,width=0.9\linewidth]{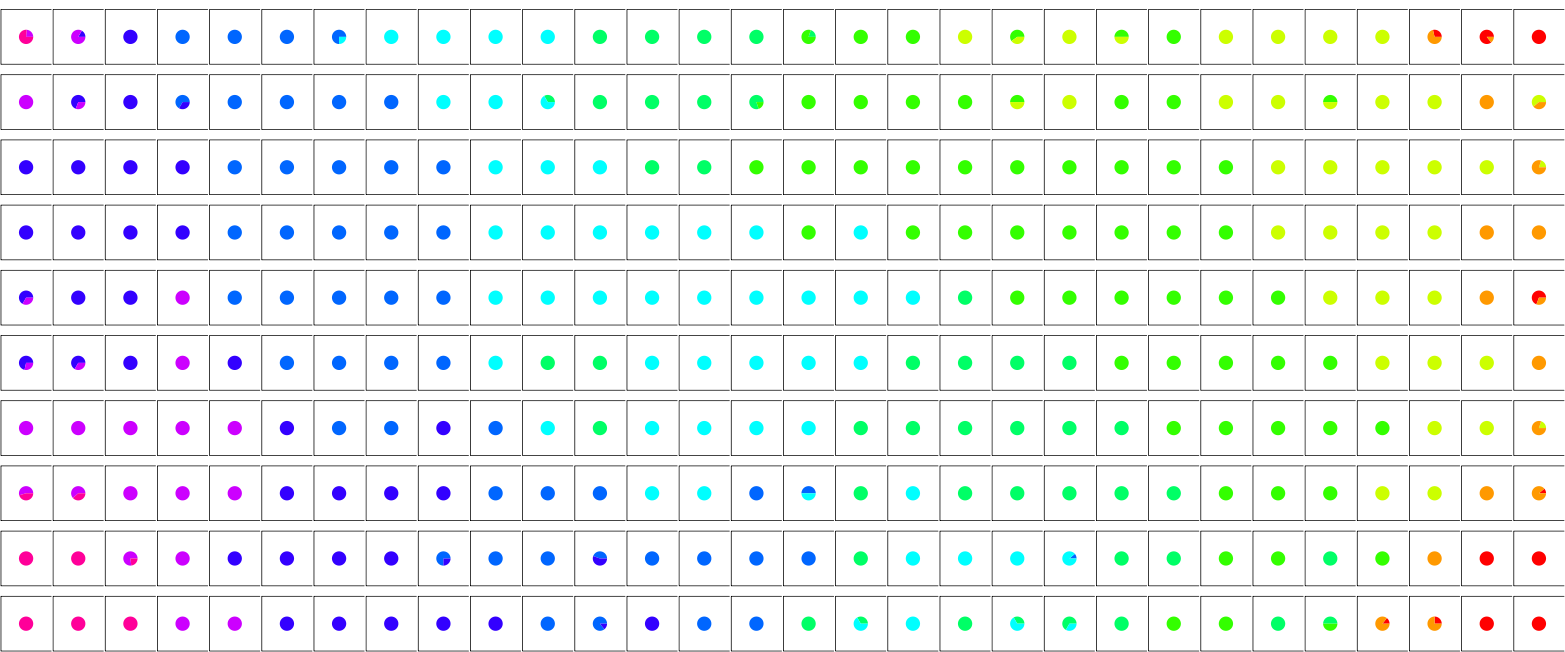}} 
    &  \fbox{\includegraphics[height=3cm,width=0.9\linewidth]{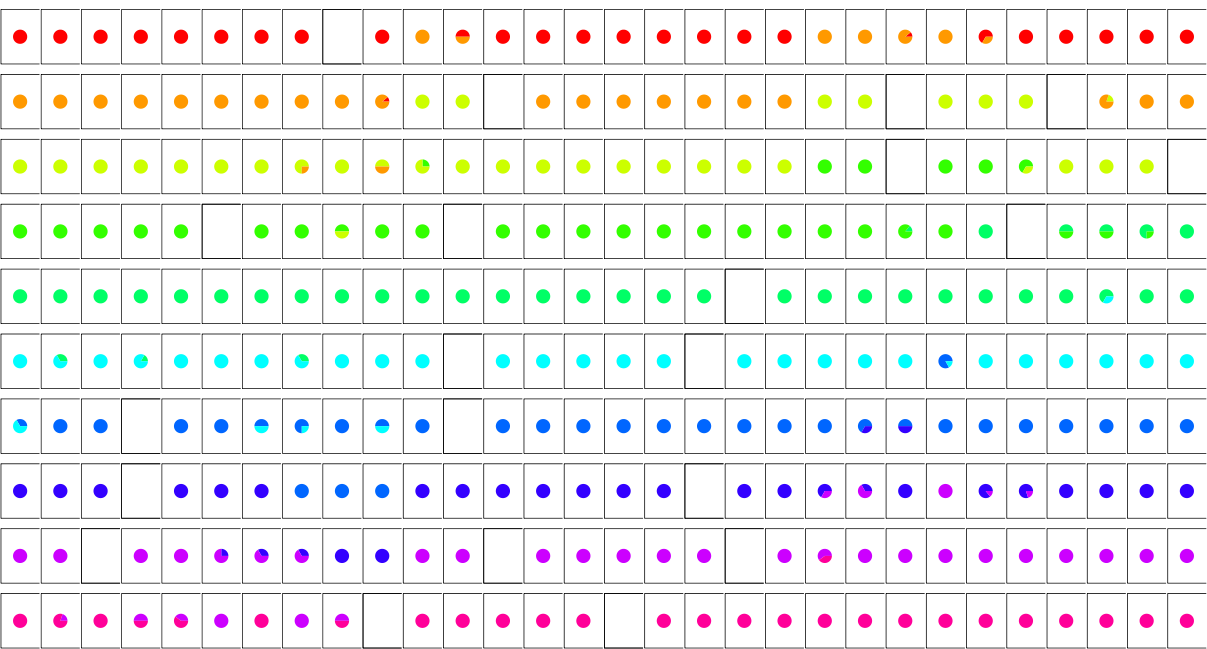}} \\
	\end{tabular}
	\caption{Unfolding the Swiss roll using self-organizing maps}
	\label{swiss-som}
\end{figure}

Let us first use a toy example to illustrate the stochastic version of relational SOM. The simulated data is the 
popular Swiss roll, a two-dimensional manifold embedded in a three-dimensional space. This example has already 
been used for illustrating the performances of Isomap
\cite{TenenbaumEtAl2000}. The data has the shape illustrated by Figure~\ref{fig::swiss}.  5~000 points were
simulated. However, since all methods
presented here work with matrices of pairwise distances, the computation times would have
been rather heavy for 5~000 points. Hence, we run the different algorithms on
1~000 points uniformly distributed
on the manifold. First, the distance matrix was computed using the
geodesic distance based on the
$K$-rule with $K=10$. Then, two types of algorithms were performed:
multidimensional scaling and self-organizing
maps. The results obtained with Isomap \cite{TenenbaumEtAl2000} are available in
Figure~\ref{fig::swiss}. As expected,
both methods succeed in unfolding the Swiss roll and the results are very similar. Next, batch median SOM and on-line relational SOM
were applied to the dissimilarity
matrix
computed with the geodesic distance. As shown in Figure~\ref{swiss-som}, the size of the map plays an important 
role in unfolding the data. For squared grids, the problem is not completely solved by either of the two 
algorithms. Nevertheless, on-line relational SOM manages to project the different scrolls of the roll into separate
regions on the map. Moreover, some empty cells highlight the roll structure, which is not completely unfolded
but rather projected without overlapping. Since squared grids appeared too heavily constrained, we also tested
rectangular grids. The results are better for both algorithms which both manage to unfold the data. However, the on-line 
version clearly outperforms the batch version.

\subsection{Amazonian butterflies}\label{astraptes}

This data set contains 465 input data and was previously used by \cite{hebert_ten_2004} to demonstrate the synergy between DNA barcoding
and morphological-diversity studies. The notion of DNA barcoding comprises a wide family of 
molecular and bioinformatics methods aimed at identifying biological specimens and assigning them to a species.
 According to the vast literature published during the past years on the topic,
two separate tasks emerge for DNA barcoding: on the one hand, assign unknown samples to known species and,
on the other hand, discover undescribed species, \cite{DeSalle2005}. The second task is usually approached with
the Neighbor Joining algorithm \cite{Saitou01071987} which constructs a tree
similar to a dendrogram. When the
sample size is large, the trees become rapidly unreadable. Moreover, they are quite sensitive to the order in 
which the input data are presented. Let us also mention that unsupervised learning and
visualization methods are used to a very limited extent by the DNA barcoding community, although the 
information they bring may be quite useful. The use of self-organizing maps may be quite helpful in visualizing 
the data and bringing out clusters or groups of clusters that may correspond to undescribed species. 

\begin{figure}[h!]
	\centering
	\begin{tabular}{p{6cm}p{6cm}}
     a) Species diversity (radius proportional to the size of the cluster) & 
     b) Distances between prototypes \\
      \fbox{\includegraphics[height=4cm,width=0.9\linewidth]{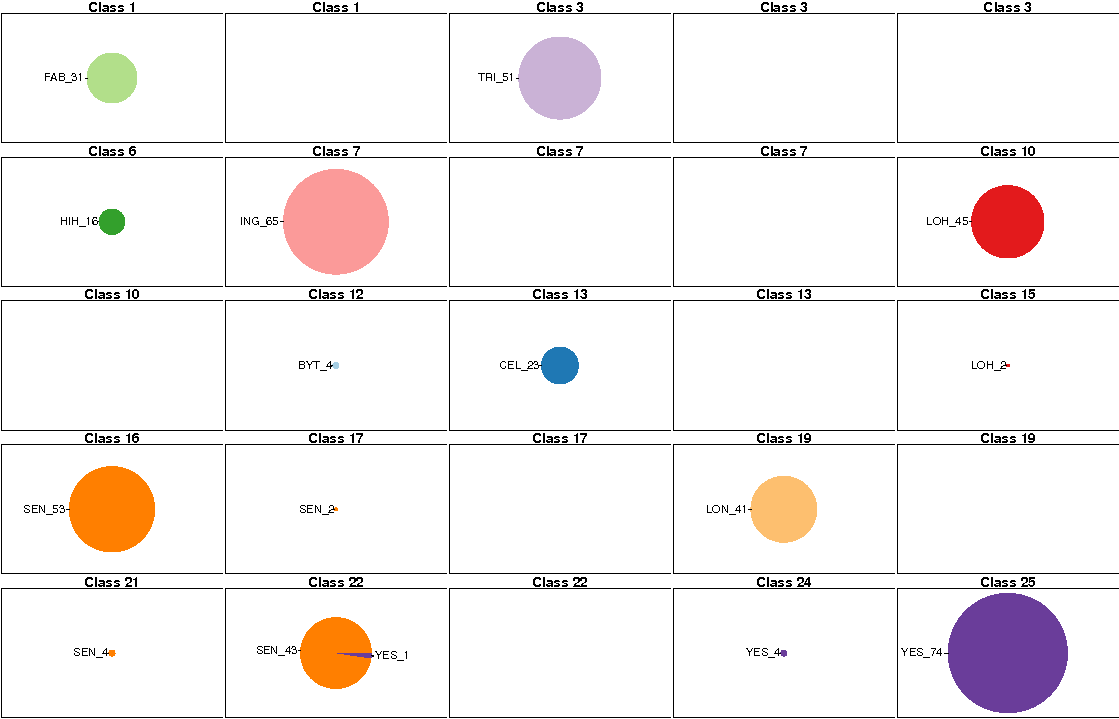}} 
    &  \fbox{\includegraphics[height=4cm,width=0.9\linewidth]{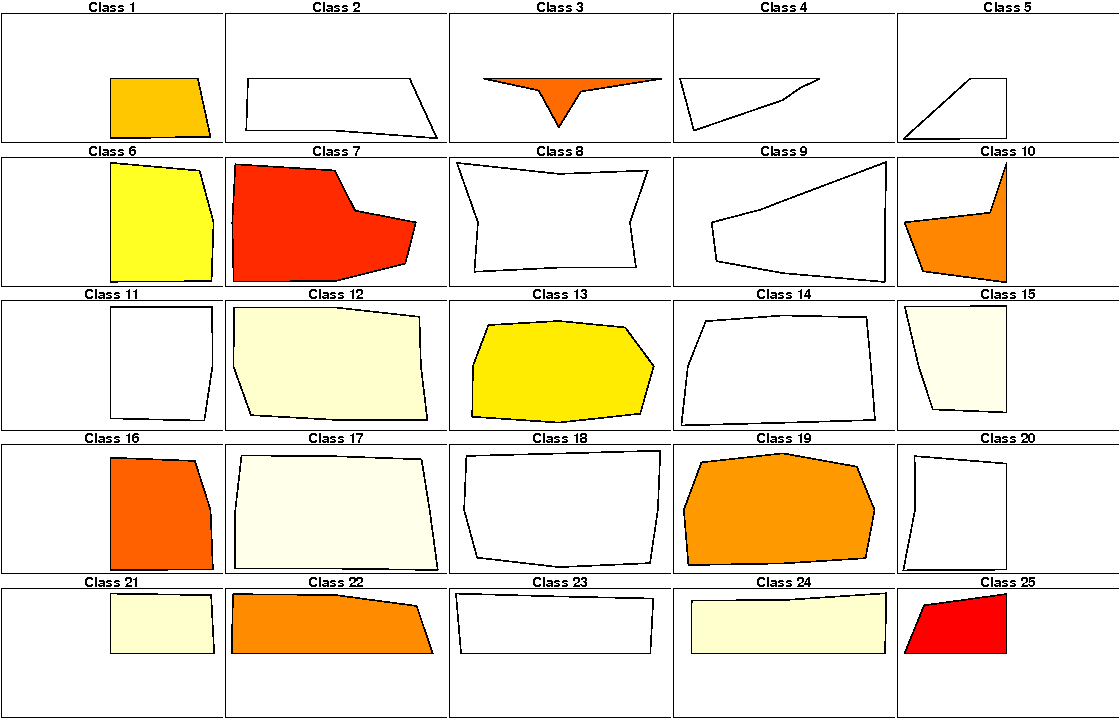}} \\
	\end{tabular}
	\caption{On-line relational SOM for Amazonian butterflies}
	\label{papillons}
\end{figure}

DNA barcoding data are composed of sequences of nucleotides, i.e. sequences of ``a'', ``c'', ``g'', ``t'' letters
in high dimension (hundreds or thousands of sites). Specific distances and dissimilarities such as the Kimura-2P 
(\cite{kimura80}) are usually computed. Hence, since the data is not Euclidean, dissimilarity-based methods
appear to be more appropriate. Recently, batch median SOM was
tested in \cite{olteanu_etal_molec} on several data sets,
amongst which the Amazonian butterflies. Although median SOM provided encouraging results, two main drawbacks emerged.
First, since the algorithm was run in batch, the organization of the map was generally poor and highly
depending on the initialization. Second, since the algorithm
calculates a prototype for each cluster among the dataset, it does not
allow for empty clusters. Thus, the existence of species or groups of species
was difficult to acknowledge. The use
of on-line relational SOM overcomes these two issues. As shown in Figure~\ref{papillons}, clusters are generally
not mixing species, while the empty cells allow detecting the main groups of species. The only mixing class
corresponds to a labeling error. Unsupervised clustering may thus be useful in addressing misidentification
issues.  In Figure~\ref{papillons}b, distances with respect to the nearest neighbors were computed for each node. 
The distance between two nodes/cells is computed as the mean dissimilarity between the observations within each class. 
A polygon is drawn within each cell with vertices proportional to the distances to its neighbors. 
If two neighbor prototypes are very close, then the corresponding vertices are very close to the edges of the two cells. 
If the distance between neighbor prototypes is very large, then the corresponding vertices are far apart, close to the center of the cells.  

\subsection{Political books}\label{polbooks}

This application uses a dataset modeled by a graph having 105 nodes. The nodes
are books about US politics published around the time of the 2004 presidential
election and sold by the on-line bookseller Amazon.com. Edges between two nodes
represent frequent co-purchasing of the two books by the same buyers. The graph
contains 441 edges and all nodes are labeled according to their political
orientation (conservative, liberal or neutral). The graph has been extracted by
Valdis Krebs and can be downloaded at
\url{http://www-personal.umich.edu/~mejn/netdata/polbooks.zip}.

\begin{figure}[h!]
	\centering
	\begin{tabular}{cc}
		\includegraphics[width=0.45\linewidth]{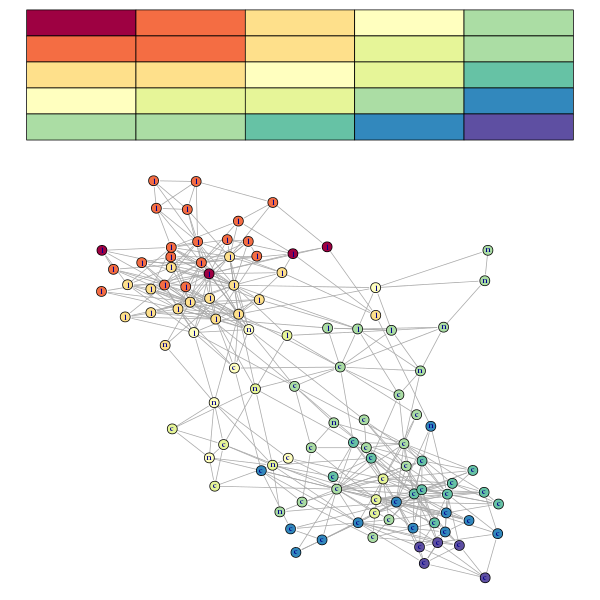}
	\end{tabular}
	\caption{``Political books'' network displayed with a force directed
placement algorithm. The nodes are labeled according to their political
orientation and are colored according to a gradient that aims at emphasizing
the distance between clusters on the grid, as represented at the top the
figure.}
	\label{fig::polbooks-graph}
\end{figure}

\begin{figure}[h!]
	\centering
	\begin{tabular}{cc}
		\includegraphics[width=0.45\linewidth]{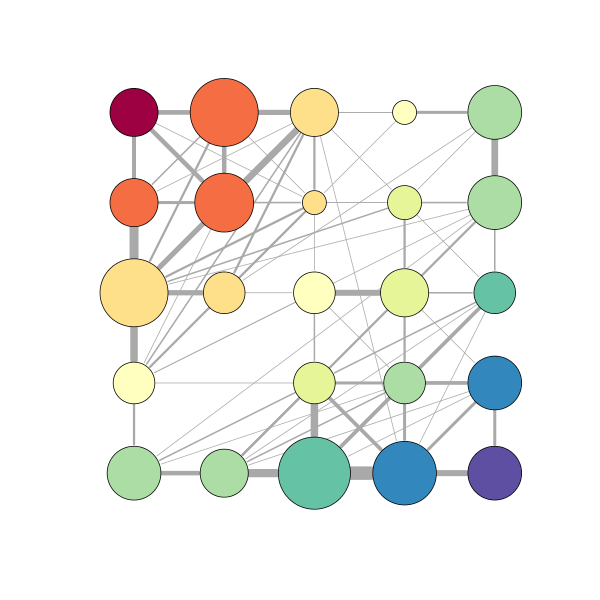} &
\includegraphics[width=0.5\linewidth]{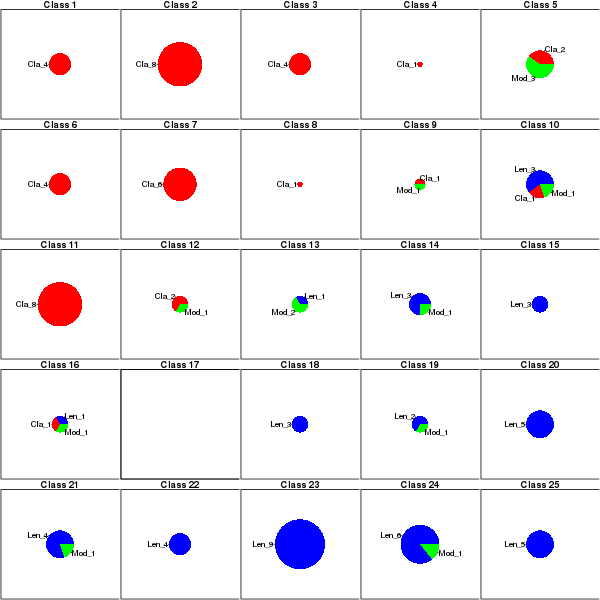}
	\end{tabular}
	\caption{Left: Simplified representation of the graph on the grid: each node
represents a cluster whose area is proportional to the number of nodes included
in it and the edges width represents the number of edges between the nodes of
the corresponding cluster. Right: Distribution of the node labels for each
neuron of the grid for the
clustering obtained with the dissimilarity based on the length of the shortest
paths. Red is for liberal books, blue for conservative books and green for
neutral books.}
	\label{fig::polbooks-projgraph}
\end{figure}

On-line relational SOM was used to cluster the nodes of the graph,
according to the length of the shortest path between two nodes, which is a
standard dissimilarity measure between nodes in a
graph. Figures~\ref{fig::polbooks-graph} and
\ref{fig::polbooks-projgraph} (left) provide two representations of the
``political books'' network: the first one is the original graph displayed with
a force directed placement algorithm, which is the one described in
\cite{fruchterman_reingold_SPE1991} and colored according to the clusters in
which the nodes are classified. The second one is a simplified
representation of the graph on the grid, where each node represents a cluster.
The colors in the first figure and the density of edges in the second one shows
that the clustering has a good organization on the grid, according to the graph
structure: groups of nodes that are densely connected are classified in the
same or in close clusters whereas groups of nodes that are not connected are
classified apart.

Additionally, Figure~\ref{fig::polbooks-projgraph} provides the
distribution of the
node labels inside each cluster for the obtained clustering (on the right hand
part of the figure).
Almost all clusters
contain books having the same political orientation. Clusters that contain
books with multiple political orientations are in the middle of the grid and
include neutral books. Hence, this clustering can give a clue on a more subtle
political orientation than the original labeling: for instance, liberal books
from cluster 12 probably have a weaker commitment that those from clusters 1 or
2.

\section{Conclusion}

An on-line version of relational SOM is introduced in this paper. It combines
the standard advantages of the stochastic version of the SOM (better
organization and faster computation) with the relational SOM that is able to
handle data described by a dissimilarity. The algorithm shows
good performances in projecting data described either by numerical variables or by
categorical variable, as well as in clustering the nodes of a graph.

\bibliographystyle{splncs03}
\bibliography{bibliototal}

\end{document}